\documentclass[twocolumn]{svjour3}          % twocolumn
\smartqed  % flush right qed marks, e.g. at end of proof

\usepackage[numbers]{natbib}
\usepackage{graphicx}

\usepackage{epstopdf}
\usepackage{mathptmx}      % use Times fonts if available on your TeX system
\usepackage[none]{hyphenat}
\usepackage{amsmath}
\usepackage{amssymb}

\usepackage{ulem}
\usepackage{subcaption}
\usepackage{float}

\usepackage[ruled]{algorithm2e}

\usepackage[usenames,dvipsnames]{color}
\usepackage[colorlinks=true]{hyperref}
\hypersetup{
  colorlinks,
  citecolor=Blue,
  linkcolor=Red,
  urlcolor=Violet}
  
\begin{document}
\sloppy
\title{Spectral video construction from RGB video: Application to Image Guided Neurosurgery}

\author{Md. Abul Hasnat \and Jussi Parkkinen \and Markku Hauta-Kasari}
\institute{Md. Abul Hasnat, \email{md-abul.hasnat@ec-lyon.fr} \at
  		   Laboratoire LIRIS, \'Ecole Centrale de Lyon, Ecully, France. \\
		   Jussi Parkkinen, \email{jussi.parkkinen@uef.fi}\\	
		   Markku Hauta-Kasari, \email{markku.hauta-kasari@uef.fi} \at
  		   School of Computing, University of Eastern Finland, Joensuu, Finland.	
}

\maketitle

\begin{abstract}
Spectral imaging has received enormous interest in the field of medical imaging modalities. It provides a powerful tool for the analysis of different organs and non-invasive tissues. Therefore, significant amount of research has been conducted to explore the possibility of using spectral imaging in biomedical applications. To observe spectral image information in real time during surgery and monitor the temporal changes in the organs and tissues is a demanding task. Available spectral imaging devices are not sufficient to accomplish this task with an acceptable spatial and spectral resolution. A solution to this problem is to estimate the spectral video from RGB video and perform visualization with the most prominent spectral bands. In this research, we propose a framework to generate neurosurgery spectral video from RGB video. A spectral estimation technique is applied on each RGB video frames. The RGB video is captured using a digital camera connected with an operational microscope dedicated to neurosurgery. A database of neurosurgery spectral images is used to collect training data and evaluate the estimation accuracy. A searching technique is used to identify the best training set. Five different spectrum estimation techniques are experimented to indentify the best method. Although this framework is established for neurosurgery spectral video generation, however, the methodology outlined here would also be applicable to other similar research.
.
\end{abstract}

\section{Introduction}
\label{sec:intro}
The objective of spectral imaging is to provide an accurate representation of the color information of an object being imaged. Spectral imaging is an imaging technnique where each pixel contains a full wavelength spectrum of all points in the image scene. Spectral image allows reproduction of color image under various illuminations, analyze images of individual spectral band and study computational color vision and color constancy \cite{shimano2007recovery}. Individual spectral band image provides better discrimination of objects of interest, segmentation, recognition and classification of objects. Spectral imaging technology has been used in various fields of applications \cite{hardeberg2001acquisition} (technology: automatic inspection, remote sensing; industry: paper, ink, painting, printing, wood, textile, agriculture, food, chemical and medicine; service: medical). Ongoing research work on exploring the possibility of its use in several other areas reveals the importance of spectral imaging.

The task of spectral imaging can be accomplished in various ways. One approach is to use different narrow band filters with a grayscale image capturing device. A filter wheel consisting of narrowband filters (interference filter systems \cite{mansouri2005optical}), an acousto-optic tunable filter (AOTF) \cite{cheng1993multispectral, morris1994imaging}, or a liquid crystal tunable filter (LCTF) \cite{morris1994imaging} are often employed in this kind of imaging systems. Therefore, in these systems an image is captured by scanning in the spectral domain. Another approach is based on spatial scanning, where the spectral information is measured line by line \cite{hyvarinen1998direct}. The main differences between the available spectral imaging systems are in accuracy, speed and scanning methods. The spectral and the spatial resolutions of the line scanning cameras (PGP based devices \cite{hyvarinen1998direct}) are usually very high and one line spectral measurement can be done rapidly. However, for capturing the whole image they are quite slow in comparison to the other imaging systems. Among the filter based devices, LCTF \cite{morris1994imaging} and AOTF \cite{cheng1993multispectral, morris1994imaging} systems are faster than interference filter systems.

An important drawback of the available spectral imaging devices is the speed of image acquisition. Despite the invent of many recent spectral imaging devices for faster acquisition, a near real time spectral image capturing device with reasonable spatial resolution is yet to become available. In order to overcome this limitation, several alternative approaches have been proposed \cite{yi2011instrument, yamaguchi2010multispectral, du2009prism, park2007multispectral}. One approach is to produce a custom designed spectral imager with identified spectral range of interest \cite{yi2011instrument}. Another similar approach \cite{yamaguchi2010multispectral} is to add spectral bands with 3-RGB bands. A different approach is to recover spectral reflectance in a scene with a conventional RGB camera using multiplex illumination \cite{park2007multispectral}. Beside these techniques, there are several computational approaches to obtain spectral reflectance with the use of conventional 3 band camera through spectrum estimation \cite{shimano2007recovery, haneishi2000system, stigell2007wiener, heikkinen2007regularized, cheung2005characterization, lopez2007selecting, connah2004comparison, imai2000comparative, imai1999spectral, morovic2006metamer, westland2012computational, solli2005color, shi2002using}. These approaches estimate spectral reflectance using various mathematical methods with the aid of prior knowledge. The aim of these estimation approaches is to reduce the cost and complexity of the image acquisition system while preserving its colorimetric and spectral accuracy. The prior knowledge includes the information about the spectra of the standard color patches or target object, corresponding response of traditional imaging device, sensor sensitivity, illumination spectra and noise.

A reasonable number of estimation techniques have been proposed in literature and usefulness of these techniques was evaluated for variety of different purposes \cite{shimano2007recovery, haneishi2000system, stigell2007wiener, heikkinen2007regularized, cheung2005characterization, lopez2007selecting, connah2004comparison, imai2000comparative, imai1999spectral, morovic2006metamer, westland2012computational, solli2005color, shi2002using}. Considering the behavior of the methods and availability of required data, these methods are categorized in \cite{shimano2007recovery, connah2004comparison}. One of the most widely used techniques is the Wiener estimation \cite{shimano2007recovery, haneishi2000system, stigell2007wiener, lopez2007selecting}, which minimizes the mean square error (MSE) between the recovered and measured spectral reflectance. In the Wiener method, estimation matrix can be computed from different set of spectra information. One set of spectral information is the reflectance spectra of training samples, the illumination spectrum and the device sensitivity \cite{shimano2007recovery, haneishi2000system}. The alternative set consists of only reflectance spectra of the training set and corresponding RGB values \cite{stigell2007wiener, lopez2007selecting}. A modification of the Wiener estimation is accomplished by a technique called the Pseudoinverse transformation method \cite{shimano2007recovery}, which is also called the least square method in \cite{lopez2007selecting}. This technique uses regression analysis \cite{murphy2012machine} between the known spectral reflectance and corresponding device response values. Another category of the methods are the linear methods where the spectral reflectance is represented as a linear combination of the orthonormal basis vectors \cite{shimano2007recovery, connah2004comparison}. Similar to the modification of the Wiener estimation method, a regression analysis is employed to modify the linear method. In this modification, a relation is established between the weight column vectors for the basis vectors and corresponding sensor responses (digital counts of the trichromatic camera). This modified method is called the \textit{Imai-Berns} method \cite{shimano2007recovery, cheung2005characterization, lopez2007selecting, imai2000comparative, imai1999spectral}. One of the major foci of the \textit{Imai-Berns} method is to determine the optimal number of channels and basis vectors for sufficient estimation. According to their suggestion, the number of channels should be the same or larger than the number of basis vectors. They tested various combinations for the number of eigenvectors and channels and found that six channels is the best compromise between accuracy and the cost of adding more sets of trichromatic channels. Shi and Healey further investigated to determine the optimal number of basis vectors \cite{shi2002using}. In their color scanner calibration experiment, they have shown that, multiple solutions exists for same device response when the number of basis vectors is more than the number of channels. In most devices three channels are used. However, in most cases three basis vectors will not be sufficient to represent accurately a large set of reflectance functions. In order to overcome this situation, Shi and Healey proposed a new method \cite{shimano2007recovery, cheung2005characterization, lopez2007selecting, shi2002using} that recovers the spectral reflectances with more basis vectors than sensors while still increasing the accuracy of the recovery.

Spectral imaging has received increasing interest in the field of medical imaging modalities \cite{falt2009extending, gebhart2006liquid, kinnunen2010optical, akbari2009hyperspectral, keereweer2011optical, antikainen2011spectral, leitner2013multi, crane2010multispectral, roblyer2008multispectral, toms2005intraoperative}. With the invent of new spectral imaging based technologies which are customized for medical imaging applications, spectral imaging has been used for imaging diabetic retinopathy \cite{falt2009extending}, imaging neurosurgical target tissues \cite{antikainen2011spectral}, determine an adequate tumor-free margin \cite{keereweer2011optical}, evaluating the degeneration of articular cartilage \cite{kinnunen2010optical}, distinguishing normal and neoplastic tissues in the brain \cite{gebhart2006liquid}, detecting cancerous tissue \cite{leitner2013multi}, detection of the sentinel lymph node in gynecologic oncology \cite{crane2010multispectral} and detection of oral neoplasia \cite{roblyer2008multispectral}. Moreover, various use of spectral imaging technology in medical field are listed in literature \cite{akbari2009hyperspectral}, such as evaluation of tissue oxygen, diagnosis of hemorrhagic shock and detection of chronic mesenteric ischemia. Spectral imaging technology has proven its capability to produce reliable information for the surgeons with sufficient sensitivity in order to discriminating different types of tissues. Previous studies have already shown that the delineation of neoplastic tissue from normal tissue can be enhanced by optical spectroscopy \cite{toms2005intraoperative} and spectral imaging devices \cite{gebhart2006liquid}. In addition to this, spectral analysis can be expected to provide useful clinical information for other purposes such as monitoring of blood flow or unwanted changes in normal tissues during surgeries.

Real time image capture and visualization for intra-operative image-guided surgery \cite{keereweer2011optical, leitner2013multi, crane2010multispectral} provides significant benefit to the surgeons. It is an immense interest from the surgeons to observe spectral images real time during surgery and monitor important information such as temporal changes in the organs and tissues as well as visual difference between various target tissues. Therefore, a spectral video system is necessary to incorporate during surgery. Keereweer et. al. \cite{keereweer2011optical} briefly discussed about the issues related to real time optical image guided surgery. They mentioned about several devices to accomplish this task where the images are considered to be taken within the range of NIR wavelengths. A multispectral real-time fluorescence imaging \cite{crane2010multispectral} is used for detecting the sentinel lymph node where the multispectral signals from different cameras are combined to yield true quantitative fluorochrome bio-distribution. Most recently, Leitner et. al. \cite{leitner2013multi} presents a multispectral video endoscopy system which comprises of 8 spectral band and 40 frames per second. They need to deal with image registration issues in order to obtain accurate spectral image of a moving organ or tissue. The limitations of the abovementioned real time systems are: very expensive, portability, limited number of spectral bands, additional processing of individual spectral band image and lower frame rate. 

Currently no spectral video capturing device exists which has sufficiently high spatial resolution and covers a wide range of spectrum. A possible solution is that, rather than using spectral video system, a RGB video camera in cooperation with a spectrum estimation technique can be used. Then spectral video can be generated from RGB video in near real time. This research work is motivated from this solution, and proposes a framework to generate spectral video from neurosurgery RGB video. The RGB video is captured using a digital camera connected to an operational microscope dedicated to neurosurgery \cite{antikainen2011spectral}. The proposed framework successfully overcomes the limitations of the existing solutions in terms of expense, portability, range and number of spectral bands and frame rate. Moreover, we believe that the framework is extendable and customizable depending on the demand of applications.

Our key contribution in this research is to propose a complete framework for generating multispectral video from RGB video. Rather than proposing device oriented solution, our proposed solution relies on computational estimation and machine learning based approach. In order to ensure best estimate of the spectra from rgb, five different estimation methods are implemented and their performance is evaluated. The best training dataset is determined to compute the desire estimation matrix which can be used for any further estimation of the neurosurgery video frames.

We discuss the methodology of spectral estimation and experimental procedure in Section \ref{sec:method}, then present the results and discuss about them in section \ref{sec:res_discuss} and finally section \ref{sec:conclusions} draws conclusions. 
\section{Methodology}
\label{sec:method}
In this section, first we present several spectral estimation techniques in Sec. \ref{ssec:sp_est_met} and then present the complete framework and experimental procedure for spectral video generation in Sec. \ref{ssec:exp_procedure}.
\subsection{Spectral estimation methods}
\label{ssec:sp_est_met}
The basic idea of spectral estimation is to compute spectral response of a pixel value consisting RGB responses from a digital camera. The estimated spectral response will be a higher dimensional representation of the color signal. With the assumption that all surfaces are Lambertian and there is no fluorescence, the device response $\rho$ of a camera system with its associative reflectance $r(\lambda)$ under an illuminant $L(\lambda)$ can be modeled by \cite{connah2004comparison}:
\begin{equation}
\rho_i = \int_\lambda S_i(\lambda) L(\lambda) r(\lambda) \; d\lambda + \delta
\label{eq:cam_syst}
\end{equation}
where, $\rho_i$ is the response for the $i^{th}$ camera channel, $S_i(\lambda)$ is the spectral sensitivity of that channel, $\delta$  represents measurement noise and $\lambda$  represents the wavelength. Most commonly there are 3 channels (R, G and B) of the camera, which means the usual value of $i$ =1, 2 or 3. Eq. (\ref{eq:cam_syst}) can be represented in terms of vectors and matrices as \cite{shimano2007recovery}:
\begin{equation}
\rho = SLr \; + \delta
\label{eq:cam_syst_vec}
\end{equation}
where, $S$ (spectral sensitivities of sensors) is an M$\times$N matrix, $L$ (spectral power distribution of an illuminant) is a N$\times$N diagonal matrix with the samples along the diagonal, $r$ is a N$\times$1 vector with the reflectance values and $\delta$ is a M$\times$1 additive noise vector. Here, $N$ represents the number of samples over the visible wavelengths, and $M$ represents the number of sensors. Now, considering the camera system model in Eq. (\ref{eq:cam_syst}) and (\ref{eq:cam_syst_vec}), our goal is to directly estimate the spectral reflectance $\hat{r}$ from the device response $\rho$. Next, we briefly present the details of several estimation methods that we experiment in this research.
\subsubsection{Wiener estimation method}
\label{sssec:wiener_est}
Let us consider that image pixels are measured with reflectance $r$. After that, the same pixel values are measured with a multichannel digital camera with device response $\rho$. We assume that the spectrum of the illumination $L$, sensitivity of the multichannel digital device $S$ and associative noise of the camera $\delta$ are already known to us. \textit{Wiener} estimation method \cite{shimano2007recovery, haneishi2000system, stigell2007wiener, lopez2007selecting} was developed on the basis of minimum mean square error (MMSE) criterion. If we recall Eq. (\ref{eq:cam_syst_vec}) and consider $Q = SL$ ($Q$ will be a matrix of dimension M$\times$N) then we can rewrite it as \cite{haneishi2000system}:
\begin{equation}
\rho = Qr + \delta
\label{eq:cam_syst_wiener}
\end{equation}

From Eq. (\ref{eq:cam_syst_wiener}), the \textit{Wiener} estimation can be obtained by operating a matrix $W$ to the device response $\rho$ as:
\begin{equation}
\hat{r} = W\rho
\label{eq:wiener_est}
\end{equation}
were, the operator $W$ is determined such that it satisfy the MMSE criterion, i.e. it minimizes the ensemble average between the original ($r$) and estimated ($\hat{r}$) spectral reflectances \cite{haneishi2000system}:
\begin{equation}
\langle \Vert r - \hat{r} \Vert^2 \rangle = \langle \Vert r - W\rho \Vert^2 \rangle \rightarrow min
\label{eq:min_est_sp_ref}
\end{equation}
where, $\langle\rangle$ denote an ensemble average. The explicit form \cite{shimano2007recovery, haneishi2000system} of the \textit{Wiener} estimation is given by:
\begin{equation}
W= R_{ss} Q^t \left(  QR_{ss} Q^t + R_{\delta\delta} \right) ^{-1}  
\label{eq:exp_wiener_form}
\end{equation}
where, $R_{ss}$ denotes the autocorrelation matrix of spectral reflectances of the learning samples and $R_{\delta\delta}$ denotes the autocorrelation matrix of measured noise. These autocorrelation matrices are computed as:
\begin{equation}
R_{ss} = \langle rr^t \rangle, \;  R_{\delta\delta} = \langle \delta\delta^t \rangle   
\label{eq:auto_corr_mat}
\end{equation}

Eq. (\ref{eq:exp_wiener_form}) computes the \textit{Wiener} estimation matrix \cite{shimano2007recovery, haneishi2000system} considering the spectral reflectance of training spectrum ($r$), spectrum of illumination ($L$) and device sensitivity information ($S$). An alternative way \cite{stigell2007wiener} to compute the estimation matrix from the spectral reflectance of training spectrum ($r$) and device response ($\rho$) as:
\begin{equation}
W = R_{r\rho} R_{\rho\rho}^{-1}
\label{eq:auto_corr_mat}
\end{equation}
where, $R_{r\rho}$ is a cross-correlation matrix between vectors $r$ and $\rho$, and $R_{\rho\rho}$ is an autocorrelation matrix of vector $\rho$. These two correlation matrices are defined as
\begin{equation}
R_{r\rho} = \langle r\rho^t \rangle,  \; R_{\rho\rho} = \langle \rho\rho^t \rangle  
\label{eq:correlation_mat}
\end{equation}

Therefore, the spectral reflectance ($\hat{r}$) can be estimated by using the estimation matrix ($W$ computed from Eq. (\ref{eq:exp_wiener_form}) or (\ref{eq:auto_corr_mat})) into the Eq. (\ref{eq:wiener_est}).
\subsubsection{Pseudoinverse estimation method}
\label{sssec:psinv_est}
The idea of pseudoinverse estimation \cite{shimano2007recovery} (also called least square \cite{lopez2007selecting}) method is to establish a direct mapping system from device response to reflectance that minimizes the least square error for a characterization or training set of known reflectance spectra with associative device responses. Let us consider, $P$ as a M$\times$k matrix that contains the sensor responses $[ \rho_1 ,\rho_2 , \ldots ,\rho_k]$ and let $R$ be a N$\times$k matrix that contains the corresponding spectral reflectances $[r_1 ,r_2 , \ldots ,r_k]$ where $k$ is the number of learning samples. The goal of pseudoinverse method \cite{shimano2007recovery, lopez2007selecting} is to compute a transformation matrix $W$ with dimension N$\times$M, that minimizes $\Vert R - WP\Vert$. The notation $\Vert . \Vert$ represents the Frobenius norm. The matrix $W$ is given by:
\begin{equation}
W=RP^{+}
\label{eq:pinv_est}
\end{equation}
where $P^+$ represents the pseudoinverse matrix of the matrix $P$. Once the transformation matrix $W$ is being computed, it can be further used to estimate the reflectance from device responses as: $\hat{r}=W\rho$ which is identical to the Eq. (\ref{eq:wiener_est}).
\subsubsection{Linear estimation method}
\label{sssec:linear_est}
The concept of \textit{Linear estimation} method \cite{shimano2007recovery, connah2004comparison} came from the notion of representing the spectral reflectances as a linear combination of basis vectors. These basis vectors are derived by applying principal component analysis \cite{jolliffe2002principal} (PCA) over the spectral reflectances. Let us consider a set of reflectance $r$ with $d$ number of samples. If we perform PCA over these reflectances then we can represent the reflectance set as:
\begin{equation}
\hat{r} = w_1 v_1+ w_2 v_2 + \ldots + w_d v_d = V\omega  
\label{eq:pca_refl}
\end{equation}

In Eq. (\ref{eq:pca_refl}), $d$ represents the number of basis vectors which is equivalent to the number of samples in the reflectance $r$, $v_i$ represent the ith basis vector and $w_i$ represents it associated weight, $V$ represents an N$\times$d basis matrix containing the first $d$ basis vectors as column vectors and $\omega$ represents a d$\times$1 column vector that contains the $d$ weights corresponding to the basis vectors. Let us rearrange the basis vectors in decreasing order based on their eigenvalues. If we recall Eq. (\ref{eq:cam_syst_wiener}) without considering the noise term and substitute reflectance from Eq. (\ref{eq:pca_refl}), then we will get \cite{shimano2007recovery}:
\begin{equation}
\rho = Q\Lambda \omega = \Lambda \omega
\label{eq:cam_wt_noise_sub_refl}
\end{equation}
where, $\Lambda = QV$ is the system matrix with dimension M$\times$d computed by multiplying $Q$ (matrix with dimension M$\times$N) and $V$ (matrix with dimension N$\times$d). If we assume that number of basis vector will be equal to the number of sensors, i.e. $M=d$, then $\Lambda$ is a square matrix. Therefore, the entries of matrix $\Lambda$ will be known. Now, using Eq. (\ref{eq:cam_wt_noise_sub_refl}) we can compute the approximate weight column vector $\hat{\omega}$ by rearranging it as:
\begin{equation}
\hat{\omega}  = \Lambda^{-1} \rho
\label{eq:approx_wt_col}
\end{equation}

Note that, if our assumption does not hold, i.e. number of basis vector is more than the number of channels, then matrix $\Lambda$ will not be a square matrix. In such situation, we have to use $\Lambda^+$ instead of $\Lambda^{-1}$. Here, $\Lambda^+$ represent the pseudoinverse \cite{ben2003generalized} of matrix $\Lambda$ and $\Lambda^{-1}$ represent direct inverse of matrix $\Lambda$.

After computing the weight column vector from device responses, the estimated reflectance can be computed (by substituting Eq. (\ref{eq:approx_wt_col}) into Eq. (\ref{eq:pca_refl})) as:
\begin{equation}
\hat{r} = V \hat{\omega}  = V\Lambda^{-1} \rho 
\label{eq:est_refl_linear_model}
\end{equation}
here, $\rho$ is the multichannel device response, $\hat{r}$ is the corresponding estimation of spectral reflectance, $V$ is the matrix of basis vectors and $\Lambda$ is the system matrix computed from $S$, $L$ and $V$ matrices.
\subsubsection{Imai-Berns method}
\label{sssec:imai_berns_est}
\textit{Imai-Berns} method \cite{shimano2007recovery, cheung2005characterization, lopez2007selecting, imai2000comparative, imai1999spectral} follows similar concept of linear method to represent the spectral reflectances. However, it modifies linear method by employing a regression analysis between the weight column vectors and corresponding sensor responses \cite{shimano2007recovery}. Here, we assume that the basis vectors are arranged in decreasing order according to their eigenvalues \cite{imai2000comparative, imai1999spectral}.

Let us consider that we have a set of learning spectral reflectances $r$. We can represent the estimated reflectances in terms of basis vectors and weights with Eq. (\ref{eq:pca_refl}). Let us consider, $B$ be a d$\times$k matrix that contains the column vectors of the weights to represent the $k$ known spectral reflectances and $P$ be an M$\times$k matrix that contains corresponding sensor response vectors of those $k$ reflectances. This method establishes a relationship between weights of the reflectances $B$ and corresponding device responses $P$ as using regression analysis as $\Vert B - DP \Vert$. Here, $D$ is a matrix with d$\times$M dimension that minimizes the Frobenius norm as:
\begin{equation}
D=BP^+
\label{eq:D_mat_ib}
\end{equation}

The matrix $D$ can be used to estimate the weights from device responses $P$ as:
\begin{equation}
\hat{\omega}=DP 
\label{eq:est_wt_ib}
\end{equation}

Finally, combining Eq. (\ref{eq:est_wt_ib}) and (\ref{eq:pca_refl}) we can estimate the spectral reflectance from device response as
\begin{equation}
\hat{r}  = VDP 
\label{eq:est_sp_ref_ib}
\end{equation}
\subsubsection{Shi-Healey method}
\label{sssec:shi_healey_est}
The basic idea of \textit{Shi-Healey} method \cite{shimano2007recovery, cheung2005characterization, lopez2007selecting, shi2002using} is to generate unique solution while considering the number basis vectors to be more than the number of channels. Shi and Healey modified the linear method by allowing more than three degrees of freedom for accurate representation of spectral reflectances \cite{shi2002using}. 

Recall Eq. (\ref{eq:pca_refl}), and consider dividing the matrix of basis vectors into two matrices such as $V_1$ with N$\times$(d-M) dimension as $[v_1, v_2, … , v_{d-M}]$ and $V_2$ with N$\times$M dimension as $[v{d-2}, v_{d-1}, v_d]$. Applying this division into Eq. (\ref{eq:cam_wt_noise_sub_refl}) and (\ref{eq:est_refl_linear_model}) the sensor response $\rho$ and estimated reflectance $\hat{r}$ can be represented as:
\begin{equation}
\rho = QV_1 \omega_1  + QV_2 \omega_1 
\label{eq:sensor_resp_sh}
\end{equation}
\begin{equation}
\hat{r} =V_1 \omega_1+ V_2 \omega_2  
\label{eq:est_ref_sh}
\end{equation}
where, $\omega_1$ and $\omega_2$ represent the corresponding weight column vectors for $V_1$ and $V_2$ matrices. From Eq. (\ref{eq:sensor_resp_sh}), $\omega_2$ can be solved in terms of $\omega_1$ as:
\begin{equation}
\omega_2  = (QV_2)^{-1} (\rho - QV_1 \omega_1)      
\label{eq:omega_2_sh}
\end{equation}

Substituting $\omega_2$ into Eq. (\ref{eq:est_ref_sh}), spectral reflectance $\hat{r}$ can be represented as:
\begin{equation}
\hat{r}=V_1 \omega_1+ V_2 (QV_2)^{-1}  (\rho - QV_1 \omega_1)      
\label{eq:sp_ref_sh}
\end{equation}

In general, there will be a set of spectral reflectances that are consistent with the linear method and that satisfy $Q\hat{r} =\rho$. Let $r_i$ be the spectral reflectance vector that satisfies both conditions above. Now, the weight column vector $\hat{\omega}_1$ is derived from Eq. (\ref{eq:sp_ref_sh}) as:
\begin{equation}
\hat{\omega}_1={ V_1 - V_2 (QV_2 )^{-1} QV_1 }^{+} {r_i - V_2 (QV_2 )^{-1} \rho} 
\label{eq:wt_col_vec_sh}
\end{equation}

Therefore the estimated spectral reflectance can be obtained by the substituting Eq. (\ref{eq:wt_col_vec_sh}) into equation (\ref{eq:sp_ref_sh}) from the condition that minimizes $\Vert \hat{r}_i - r_i\Vert$.

An important characteristic of this method is that estimation time is proportional to the number of training reflectance spectra. For a large number of training spectra, this method is very slow because for every single pixel it has to check all the training spectra to find the best one as estimated spectra. In addition to this, for every single sensor response value in comparison with a particular training spectrum, it is necessary to investigate the correct number of basis vector which produce best estimation.

Different method requires different set of prior information. The common information required by all the methods are a set of spectral reflectance which is considered as training set. The prior knowledge required to estimate spectral reflectance by different methods discussed above is summarized in Table \ref{tab:prior_know_sp_est_methods} \cite{shimano2007recovery}.
\begin{table*}[]
\centering
\caption{Requirements of prior knowledge for each method \cite{shimano2007recovery}.}
\label{tab:prior_know_sp_est_methods}
\begin{tabular}{|l|c|c|c|c|}
\hline
{\bf Method Name}       & {\bf Sensitivities} & {\bf Illumination} & {\bf Reflectance} & {\bf RGB Values} \\ \hline
{\bf \it Wiener}        & Yes / No            & Yes / No           & Yes               & No / Yes         \\ \hline
{\bf \it Pseudoinverse} & No                  & No                 & Yes               & Yes              \\ \hline
{\bf \it Linear}        & Yes                 & Yes                & Yes               & No               \\ \hline
{\bf \it Imai-Berns}    & No                  & No                 & Yes               & Yes              \\ \hline
{\bf \it Shi-Healey}    & Yes                 & Yes                & Yes               & No               \\ \hline
\end{tabular}
\end{table*}

Among the spectrum estimation methods discussed here, Wiener \cite{stigell2007wiener}, Pseudoinverse \cite{shimano2007recovery} and Imai-Berns \cite{imai2000comparative} methods use linear regression analysis. In literature \cite{shimano2007recovery, stigell2007wiener, heikkinen2007regularized}, it is suggested that, use of nonlinear regression \cite{hardeberg2001acquisition} with higher order (multivariate) polynomial may improve the performance of estimation. Therefore, this research considers both linear and nonlinear regression analysis for spectral estimation experiments using Wiener \cite{stigell2007wiener}, Pseudoinverse \cite{shimano2007recovery} and Imai-Berns \cite{imai2000comparative} method.

\subsection{Spectral video generation}
\label{ssec:exp_procedure}
\begin{figure}
\centering
\includegraphics[scale=0.64]{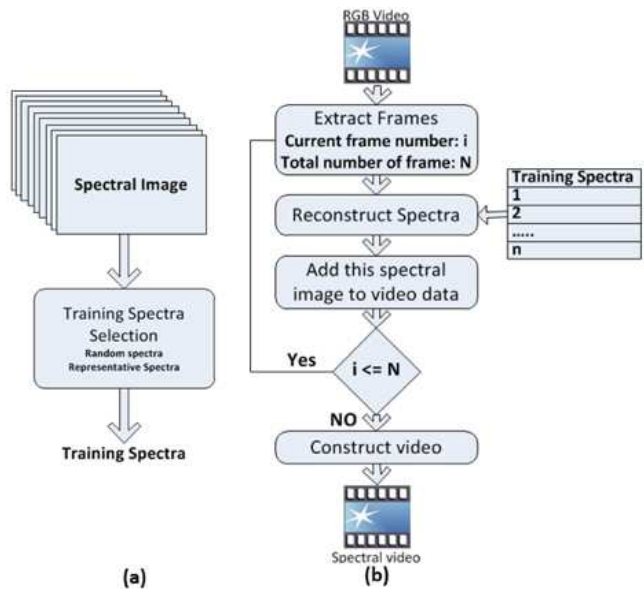} 
\caption{Block diagram of spectral video generator. (a) Training data collection (b) Spectral video generator.}
\label{fig:block_diagram}
\end{figure}

The aim of this research is to establish a framework which generates spectral video from neurosurgery RGB video. The complete system is called spectral video generation which is decomposed into two subtasks: (a) training spectra selection and (b) spectral video construction. Training spectra are considered as prerequisite knowledge for spectral estimation. Block diagram of spectral video generator is illustrated in Figure \ref{fig:block_diagram}. Initially, training spectra are collected randomly from spectral images. Then representative training set is determined (Figure \ref{fig:block_diagram}(a)) by a searching method. Figure \ref{fig:block_diagram}(b) illustrates the process to construct spectral video with the aid of prior information. These two subtasks are briefly discussed below.

Training spectra are collected from a database of spectral neurosurgery images. The database consists of 34 spectral images which were captured during surgery of 6 patients. Each spectral image was captured within the wavelength range of 420 nm to 720 nm at 10 nm interval. A spectral imaging device consisting of a camera (monochrome camera with \textit{Sony ICX414} sensor), tunable filter (\textit{VariSpecVIS VIS-10} Liquid Crystal Tunable Filter) and a focusing lens (70 mm focal length lens) was used to capture the spectral images. Since the brain tissues are moving with heart beat and breathing, each wavelength layer is slightly shifted during spectral image capture. Therefore, images must be registered accurately to achieve correct spectral responses. Among the 34 spectral images, 17 images were finally selected according to registration accuracy. Moreover, after registration few of the images contain noises and unexpected color particularly in the border. These images are manually corrected by cropping out the noisy regions. The spectral images are illustrated in Appendix \ref{append_exp_data}. RGB images corresponding to the spectral images were not captured through the digital video camera. However, for the purpose of estimation it is obvious to have a RGB color version of each spectral image. Therefore, the color images were generated from the spectral images considering standard illuminant \textit{D65} and CIE color matching functions. We consider these color images as training RGB images.

In a practical condition of spectral estimation, the training spectra are collected from pre-specified target points of the objects in image or from standard color patches (from calibration chart). In this research, the target points are not specified. Therefore, the training spectra are collected by choosing certain percentage (1\%, 5\%, 10\%, 20\% and 50\%) of random pixels from each image. As a consequence, 5 sets of training spectra are available from each spectral image. Corresponding RGB values of the selected training spectra are also collected from RGB versions of the spectral images. The assumption of this random selection is that, spectra from all possible objects in the images will be selected. In this random selection process, total 85 (17$\times$5) different set of training spectra are collected for training. However, the final goal is to select or construct only one set of training spectra which will be considered as the representative set. The representative training set is identified by observing its performance in spectrum estimation. In this research, we consider the root mean square error (RMSE) as a measure of evaluation for the training sets. RMSE is computed between original spectral image and its estimated version. The representative set of spectra may not same for all estimation methods. Therefore, the searching procedure must be applied to each estimation method separately. Steps required to search representative set of spectra are as follows:
\begin{itemize}
\item \textbf{Step 1:} For each RGB image, estimate spectral image (Sec. \ref{ssec:sp_est_met}) using each set of training spectra. Compute RMSE (Eq. (\ref{eq:rmse_calc})) and select the set with lowest RMSE.
\item \textbf{Step 2:} Using each one of the training sets (obtained from step 1), estimate all spectral images. Compute average RMSE value. Then, sort them in increasing order.
\item \textbf{Step 3:} Choose the first five sets and generate all possible combination. With these combinations, follow the procedure in step 2. Then, select the best training set according to lowest average RMSE value.
\end{itemize}

It is observed from the training spectra that, few of the spectra has quite unusual shapes and contains several peaks. These spectra are signifying the presence of highlight in the spectral image. In a neurosurgery spectral image, the highlights are present due to the liquid materials (blood, water) that reflects (specular reflection) the illuminant at certain extent. It is also interesting to note that most of the spectra collected from different objects have high intensity at the red wavelength range. The reason for this is the absence of blue and green colored objects in the collected spectral images. Another important property of the spectra is that they are not smooth.

The spectral video is generated from RGB video. At the beginning of the process the frames are sequentially extracted as RGB image from the video. A video frame is a RGB image which is first corrected if necessary. For example, for several estimation methods the RGB values need to be normalized within the range between 0 and 1. This correction is a part of preprocessing. After the preprocessing is completed, the spectral image is estimated (Sec. \ref{ssec:sp_est_met}) from the RGB image with the aid of necessary prior information and particular estimation method. Then, the estimated spectral image is added to target spectral video. This process is continued until the last video frame is reached. Temporal aspect of video processing is maintained either by including wait functionality or by skipping few frames based on inter-frame similarity. The results of spectral estimation techniques are presented in the results and discussion section.

\section{Results and discussion}
\label{sec:res_discuss}
This section evaluates the experimental performance of the spectrum estimation methods. These techniques are applied to the neurosurgery spectral image database (Appendix 1). Each spectral image has 31 spectral bands (spectral range 420 nm to 720 nm with 10 nm sampling interval).
\subsection{Spectrum estimation}
\label{ssec:spec_est}
Five spectral estimation methods (described in Sec. \ref{ssec:sp_est_met}) are experimented to estimate 17 neurosurgery spectral images from RGB images. Accuracy of estimation is evaluated by computing two spectral metric such as root mean square error (RMSE), goodness of fit coefficient (GFC) \cite{lopez2007selecting} and a colorimetric metric CIELAB ($\Delta E_ab$). The metrics are computed using the actual and estimated spectral images. With RMSE (Eq. \ref{eq:rmse_calc}) and $\Delta E_{ab}$ (Eq. \ref{eq:deab_calc}), perfect match is obtained when the value of these metrics are 0. Unlike these, in GFC (Eq. (\ref{eq:gfc_calc})) where the range of value is between 0 and 1, perfect match is obtained when the value is 1. In this research, we consider the RMSE value as the principal metric for evaluation. Additionally, we observe the values of GFC and $\Delta E_{ab}$ for further evaluation if necessary.
\begin{equation}
RMSE = \sqrt{\frac{1}{n}\sum_{j=1}^{n}\left (  r(\lambda_j) - \hat{r}(\lambda_j) \right )^2}
\label{eq:rmse_calc}
\end{equation}
\begin{equation}
GFC = \frac{\left | \sum_{j=1}^{n} r(\lambda_j) \hat{r}(\lambda_j) \right |}{ \big | \sum_{j=1}^{n} \left [ r(\lambda_j) \right ]^2 \big |^{1/2} \; \big | \sum_{j=1}^{n} \left [ \hat{r}(\lambda_j) \right ]^2 \big |^{1/2} }
\label{eq:gfc_calc}
\end{equation}
\begin{equation}
\Delta E_{ab} = \sqrt{\Delta L^{*2} + \Delta a^2 + \Delta b^2 }
\label{eq:deab_calc}
\end{equation}

In Eq. (\ref{eq:rmse_calc}) and (\ref{eq:gfc_calc}), $r$ represents the actual spectra and $\hat{r}$ represents the estimated spectra. In Eq. \ref{eq:deab_calc}, the value of $\Delta L^*$, $\Delta a$ and $\Delta b$ are calculated by first converting the spectra to XYZ values and then converting XYZ to Lab. The XYZ value of standard illuminant \textit{D65} is used as a white reference in XYZ to Lab conversion.

In the evaluation process, for each method RMSE is computed for estimating 17 images using 5 set of training spectra. Therefore there are total 85 (5$\times$17) RMSE values obtained. Then the average of these 85 RMSE values is calculated. This average RMSE is considered for evaluating the estimation methods (Sec. \ref{ssec:sp_est_met}). The method that produces the lowest average RMSE is considered as the best one. Table \ref{tab:accuracy_methods} illustrates the performance of the methods in terms of average RMSE, GFC and $\Delta E_{ab}$. We consider 3 basis vectors in \textit{linear} and \textit{Imai-Berns} methods.

\begin{table*}[]
\centering
\caption{Accuracy (average RMSE, average GFC and average $\Delta E_{ab}$) of the methods.}
\label{tab:accuracy_methods}
\begin{tabular}{|l|c|c|c|c|c|}
\hline
                   & {\bf \it Wiener} & {\bf \it Pseudo inverse} & {\bf \it Linear} & {\bf \it Imai-Berns} & {\bf \it Shi-Healey} \\ \hline
{\bf Average RMSE} & 0.029            & 0.029                    & 0.177            & 0.036                & 0.082                \\ \hline
{\bf Average GFC}  & 0.989            & 0.989                    & 0.911            & 0.972                & 0.952                \\ \hline
{\bf Average $\Delta E_{ab}$}      & 7.177            & 7.177                    & 22.130           & 12.49                & 15.623               \\ \hline
\end{tabular}
\end{table*}

From the results listed in Table \ref{tab:accuracy_methods}, it is clear that the \textit{Wiener} \cite{stigell2007wiener} and the \textit{Pseudoinverse} \cite{shimano2007recovery} methods are producing similar estimation accuracy and they are significantly different than the \textit{Linear} \cite{connah2004comparison} and the \textit{Shi-Healey} \cite{shi2002using} methods. Accuracy produces by the \textit{Imai-Berns} \cite{imai1999spectral} method is nearly close to the \textit{Wiener} and the \textit{Pseudoinverse} methods. From different metrics of accuracy measurement it is clear that when RMSE decreases then color difference ($\Delta E_{ab}$) decreases and GFC increases. This reveals that if RMSE is low, then the color difference is low and therefore the estimation accuracy is high. Because of a good agreement between the metrics, we consider only RMSE for further evaluation. 
\begin{figure*}
\centering
\includegraphics[scale=0.45]{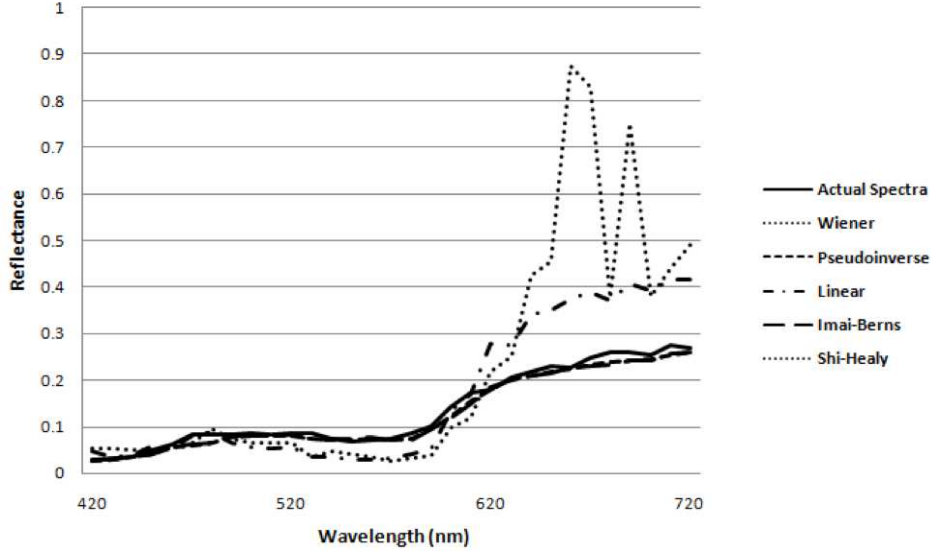} 
\caption{Plot of a recovered spectral reflectance. Comparison between actual spectra and estimated spectra from all methods.}
\label{fig:sp_refl_illustrate}
\end{figure*}

In order to compare the estimation performance visually, we consider a pixel from a vein of the brain surface for illustration. Table \ref{tab:acc_req_time} illustrates the performance of the methods (RMSE) for estimating the spectra of the selected pixel in terms of accuracy and speed. From the RMSE value of Table \ref{tab:acc_req_time}, it is clear that the \textit{Wiener} \cite{stigell2007wiener}, \textit{Pseudoinverse} \cite{shimano2007recovery} and \textit{Imai-Berns} \cite{imai1999spectral} methods outperforms the other two methods. Figure \ref{fig:sp_refl_illustrate} illustrates the recovered spectra from different methods in comparison to actual spectra. The plots clearly support the results obtained in the Table \ref{tab:acc_req_time}. In addition to the performance of estimation, the estimation time is also taken into account. Required amount of time for a particular estimation task is listed in the second row of Table \ref{tab:acc_req_time}. It is observed that estimation using the \textit{Pseudoinverse} method is faster than any other methods. Therefore, in this research we consider the \textit{Pseudoinverse} method for further experiments.
\begin{table*}[]
\centering
\caption{Accuracy and required time of the methods for estimating a spectrum of a particular pixel.}
\label{tab:acc_req_time}
\begin{tabular}{|l|c|c|c|c|c|}
\hline
                 & {\bf \it Wiener} & {\bf \it Pseudo inverse} & {\bf \it Linear} & {\bf \it Imai-Berns} & {\bf \it Shi-Healey} \\ \hline
{\bf RMSE}       & 0.0094           & 0.0094                   & 0.06             & 0.01                 & 0.036                \\ \hline
{\bf Time (sec)} & 0.031            & 0.027                    & 0.069            & 0.093                & 8.25                 \\ \hline
\end{tabular}
\end{table*}
\begin{table*}[]
\centering
\caption{Accuracy (in terms of average RMSE) of Pseudoinverse method with higher order polynomials and different number of terms.}
\label{tab:high_order_poly}
\begin{tabular}{|l|c|}
\hline
\multicolumn{1}{|c|}{{\bf Combination of polynomials and terms}} & {\bf Avg. RMSE} \\ \hline
$R, G, B, RG, GB, BR$                                              & 0.019           \\ \hline
$R, G, B, R^2, G^2, B^2$                                              & 0.018           \\ \hline
$R, G, B, R^3, G^3, B^3$                                              & 0.019           \\ \hline
$R, G, B, RG^2, GB^2, BR^2$                                           & 0.02            \\ \hline
$R, G, B, RG, GB, BR, R^2 G^2 B^2$                                    & 0.018           \\ \hline
$R, G, B, RG, BG, BR, RG^2, GB^2, BR^2$                               & 0.018           \\ \hline
$R, G, B, RG, BG, RB, R^2, G^2, B^2, RG^2, GB^2, BR^2$                   & 0.017           \\ \hline
$R, G, B, RG, BG, RB, R^2, G^2, B^2, R^2G^2, G^2B^2, B^2R^2$                & 0.017           \\ \hline
\end{tabular}
\end{table*}

In order to further increase the accuracy of estimation using the \textit{Pseudoinverse} \cite{shimano2007recovery} method, nonlinear regression analysis \cite{murphy2012machine} have been experimented. Accuracy (in terms of average RMSE) of the combinations among different higher order polynomials in cooperation with different number of terms is presented in Table \ref{tab:high_order_poly}. For each combination, average RMSE value from 85 training sets is computed as a measure of accuracy. It is observed that the accuracy increases (average RMSE decreased) with certain combinations when the number of terms is increased. However, it is also important to note that computation time is proportional to the addition of terms. Therefore, there is a tradeoff between number of terms and computation time which should be considered depending on the demand of application. In this research, we choose the combination $[R, G, B, R^2, G^2, B^2]$ for neurosurgery video estimation since it has reasonable balance between computation time and accuracy.

For evaluating the accuracy of different estimation methods, the strategy is to collect five sets of training spectra from a spectral image, and then use these training sets to experiment different methods. Finally the overall accuracy of a method is computed from the average of all these (5 sets per image $\times$ 17 spectral image = 85 sets of training spectra) accuracies. That means, each spectral image is estimated using the training samples collected from itself. However, in a practical application it is not feasible to determine a training dataset first and then estimate a particular spectral image. Therefore, a common dataset is necessary to estimate spectral image from any RGB image. We applied an algorithm (presented in section \ref{ssec:exp_procedure}) to identify the best training dataset (representative set of spectra) and observed the performance. Table \ref{tab:rmse_databases} presents the accuracy (in terms of average RMSE) of five best training datasets which are used to estimate all test spectral images. From the obtained accuracy it is observed that, there is no significant difference between these five datasets. For the purpose of selecting the representative set of spectra, we select 5\% random spectra collected from image number 2.

\begin{table}[]
\centering
\caption{Average RMSE of common training datasets to estimate all test images. Numbers indicate the percentage of data which has been collected from training images and numbers within ‘[]’ indicate the images which were combined.}
\label{tab:rmse_databases}
\begin{tabular}{|l|c|}
\hline
\multicolumn{1}{|c|}{{\bf \it Dataset}} & {\bf \it Avg. RMSE} \\ \hline
5\%, {[}2{]}                            & 0.026               \\ \hline
20\%, {[}2{]}                           & 0.027               \\ \hline
5\%, {[}2, 12, 17{]}                    & 0.03                \\ \hline
5\%, {[}2, 7, 12, 17{]}                 & 0.028               \\ \hline
5\%, {[}All{]}                          & 0.027               \\ \hline
\end{tabular}
\end{table}

In order to observe the estimation result in the image, a RGB image (single video frame) is extracted from a video. Then the corresponding spectral image is computed by the \textit{Pseudoinverse} \cite{shimano2007recovery} method using the training set. After that, RGB image (transformed RGB image) is computed from the estimated spectral image. The color reproduction error between the original and transformed RGB image is computed using \textit{S-CIELAB} \cite{zhang1997spatial} measure. The performance of the \textit{Pseudoinverse} method is evaluated using 3 terms/variables and 6 terms/variables. Corresponding \textit{S-CIELAB} values are presented in table \ref{tab:comp_scielab}. From the \textit{S-CIELAB} values it is observed that, performance of 6 terms estimation outperforms the performance of 3 terms estimation.
\begin{table}[]
\centering
\caption{Comparison (using S-CIELAB values)  between original and transformed RGB images. Entry ‘X’ means not experimented.}
\label{tab:comp_scielab}
\begin{tabular}{|c|c|c|c|c|}
\hline
\multicolumn{1}{|l|}{{\bf \it Number of terms}} & \multicolumn{1}{l|}{{\bf \it Wiener}} & \multicolumn{1}{l|}{{\bf \it Pseudoinverse}} & \multicolumn{1}{l|}{{\bf \it Linear}} & \multicolumn{1}{l|}{{\bf \it Imai-Berns}} \\ \hline
3                                               & 7.73                                  & 7.73                                         & 23.45                                 & 7.76                                      \\ \hline
6                                               & 0.95                                  & 0.95                                         & X                                     & 1.29                                      \\ \hline
\end{tabular}
\end{table}

\begin{figure*}
\centering
\includegraphics[scale=0.5]{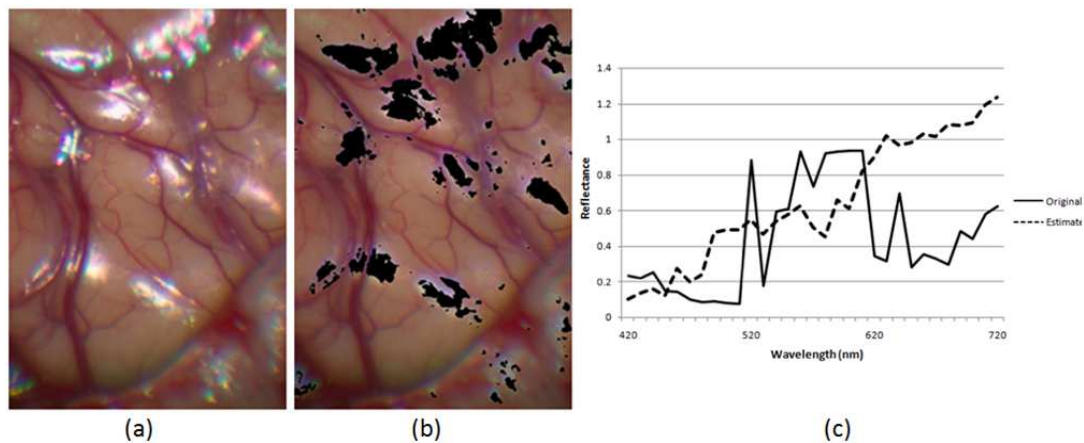} 
\caption{(a) Original image. (b) black regions indicate the pixels in image that causes high estimation error. (c) Plot of an estimated spectrum of a highlighted pixel (RMSE value is 0.43).}
\label{fig:error_img_analysis}
\end{figure*}

We further investigated the error rate produced by the \textit{Pseudoinverse} method. In order to identify the spectrum which produces large error, a threshold value is set as: 
\begin{equation}
Threshold = 2*avg\,RMSE 
\label{eq:th_large_error}
\end{equation}
Based on this threshold value we identified the pixels in the image that causes high estimation error rate. The right image (b) in figure \ref{fig:error_img_analysis} indicates the regions of the pixels where the estimated error rate is above the threshold value. It is observed that, the estimation method is giving high error rate in estimating the highlighted spectra presents in the image. In the neurosurgery spectral image, these highlights appear due to the presence of liquid materials (blood, water) that reflects the illuminant at certain extent. An example of an estimated highlight spectra (with RMSE 0.43) using the \textit{Pseudoinverse} \cite{shimano2007recovery} method is illustrated in figure \ref{fig:error_img_analysis}(c). From an experiment on the neurosurgery spectral image database with the threshold value (Eq. \ref{eq:th_large_error}), it is observed that on an average 3\% pixels are containing highlight in an image and causes increase of 0.003 average RMSE. However, this analysis does not reveal the actual effect of estimation where the highlight is present significantly.

It is observed that, if the average RMSE is computed without the highlight then it decreases from 0.26 to 0.19. For example, a particular spectral image that contains 4\% highlighted pixel, increases average RMSE value 0.024. Therefore the error rate (RMSE) is proportional to the amount of highlighted pixels. We further investigated particularly on estimating the highlighted spectra. In an analysis of RMSE values for different type of spectra, it is found that when the average RMSE value in a spectral image is 0.026, the average RMSE for non highlighted pixels are 0.019 and highlighted pixels are 0.11. Therefore, it is observed that the average RMSE value for estimating only the highlight is significantly large compare to non highlighted spectra. Therefore, more analysis and experiments with different methods are necessary to reduce the estimation error rate for highlight spectra estimation. We do not consider the task of highlight estimation in the scope of this research.
\section{Conclusions}
\label{sec:conclusions}
This research presents a framework for generating and displaying near real time neurosurgery spectral video. Each frame in the spectral video is generated using a spectrum estimation method. Five different estimation methods are examined for estimation. The estimation matrix in Wiener method is computed using the reflectance and corresponding RGB values. In the Linear and Shi-Healey methods, standard illuminant D65 and CIE color matching function are used as illumination and device sensitivity information. Pseudoinverse and Wiener estimation methods are most accurate. It is found that second order polynomial with six terms is the best choice in terms of accuracy and computational expense. Accuracy of Imai-Berns method is nearly close to Pseudoinverse and Wiener methods. Linear estimation method does not provide an acceptable accuracy. Shi-Healey method is very slow since number of training spectra considered in this research is considerably large. Therefore, in practical application of video estimation, Shi-Healey method is not a good choice with large training set. In order to collect training spectra, the most common practice is to take the spectral image of a calibration color chart along with the target objects. However in a real surgical environment, it is not possible to place a calibration color chart and capture spectral image of the chart during surgery. Therefore, in this research the training spectra are not collected from standard target patches or recommended color charts. An alternative approach is proposed to collect training spectra. In this approach, a searching technique is applied to find out the best training set from a collection of randomly selected set of spectrum. Temporal aspect in video processing is considered. Compromise between frame rate and speed of estimation is suggested when the processor speed is not sufficient to generate near real time video. 

The experimental programs in this research are written in MATLAB and the demo application is developed in C++ with OpenCV library. The methodology outlined here would also be just as applicable to other research area.

\appendix
\section{Experimental Data}
\label{append_exp_data}
\begin{figure}[h]
\centering
\includegraphics[scale=0.27]{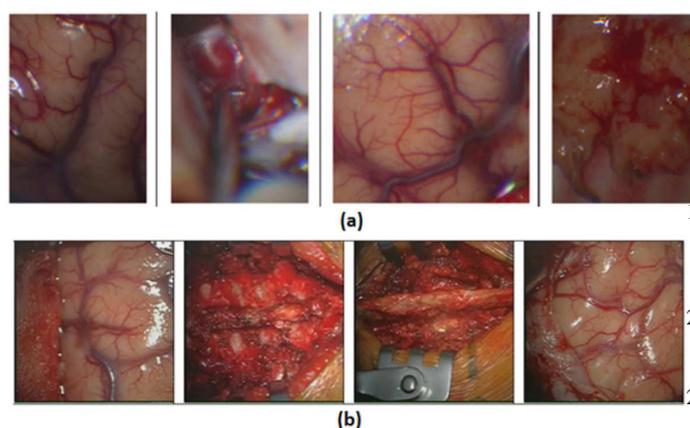} 
\caption{(a) Several representative spectral images from the database of spectral neurosurgery image. (b) Example of a frame from the four RGB videos.}
\label{fig:appnx_exp_data}
\end{figure}

Figure \ref{fig:appnx_exp_data}(a) illustrates several converted (considering standard illuminant \textit{D65} and CIE color matching function) color images from the selected spectral images. Four RGB videos (captured during surgery of different patients) are available for this research. These videos were captured using a frame grabber. Each video was captured at 720$\times$576 spatial resolution and 25 frames per second. Figure \ref{fig:appnx_exp_data}(b) illustrates an example of a frame taken from each RGB video.
\bibliographystyle{spbasic}      % basic style, author-year citations
\bibliography{sp_est_bib}
\end{document}